\documentclass[letterpaper]{article}
\providecommand{\NewCommandCopy}[2]{\let#1#2}
\usepackage[preprint]{aaai2027}
\usepackage[hyphens]{url}
\usepackage{graphicx}
\usepackage{natbib}
\usepackage{caption}
\usepackage{booktabs}
\usepackage{amsmath}
\usepackage{algorithm}
\title{EviBack: Search-Agent Reinforcement Learning via Evidence-Constrained Teacher Backoff}
\author{Xiao Ma\textsuperscript{*}, Zhiquan Hu\textsuperscript{*}, Yi Wei, Chenchen Zhao,\\
Yijun Chen, Jicheng Zhao, Yuming Li\textsuperscript{\textdagger}, Chuang Dai}
\affiliations{NEXTAI Research Institute, Chery Group\\
Project page: \url{https://chery-nextai.github.io/eviback/}.}

\begin{document}
\maketitle
\begingroup
\renewcommand{\thefootnote}{*}
\footnotetext{These authors contributed equally.}
\renewcommand{\thefootnote}{\textdagger}
\footnotetext{Corresponding author.}
\endgroup

\begin{abstract}
Search agents address a key limitation of large language models by retrieving up-to-date and proprietary information beyond their parametric knowledge. Reinforcement learning further enables search agents to learn multi-turn search from outcome rewards. However, supervision based solely on final outcome rewards leads to sparse learning signals and overlooks valuable search behaviors. We present EviBack, an evidence-constrained Teacher backoff that uses a Teacher model to provide auxiliary process supervision for sparse-reward search training, decoupled from agent's verifiable outcome rewards. By decoupling process supervision from final-answer evaluation, EviBack preserves learning signals from valuable search trajectories even when the final answer is incorrect. To make this practical, we further introduce E2E-APE, an end-to-end automatic prompt engineering pipeline that optimizes judge prompts under explicitly specified constraints and freezes a stable prompt policy for process supervision. Across seven open-domain QA benchmarks and three Qwen3 scales, EviBack improves F1 over Search-R1 and raises both single- and multi-hop macro F1, and yields more efficient, better-controlled search behavior.
\end{abstract}

\section{Introduction}

Large language models (LLMs) have demonstrated strong capabilities in knowledge-intensive question answering, text understanding, and general task solving~\cite{openai2023gpt4,yang2025qwen3,zhao2023survey}. However, the parametric knowledge of LLMs is static, making it difficult for them to perceive recent events, dynamic web content, and private enterprise data in a timely manner. When knowledge is missing, LLMs may also produce factual errors or hallucinations~\cite{ji2023survey,huang2023survey,singh2026agenticrag}. Retrieval-Augmented Generation (RAG) mitigates knowledge staleness and knowledge absence by introducing external knowledge during inference, providing the model with up-to-date or domain-relevant evidence~\cite{lewis2020rag,gao2023retrieval,singh2026agenticrag}.

Recent studies extend RAG from a fixed retrieval pipeline to a dynamic search process, where an LLM is trained as a search agent that alternates among reasoning, query generation, and tool invocation, and updates subsequent reasoning based on the returned evidence~\cite{yao2023react,asai2024selfrag,jin2025searchr1,chen2025research,li2025webthinker}. Under this paradigm, search is no longer a one-shot operation before generation, but a trajectory consisting of multi-turn querying, evidence reading, and reasoning updates. GRPO has become an important reinforcement learning method for training reasoning and search agents, since it does not require an additional value model and optimizes generated results through relative advantage estimation within sampled groups~\cite{shao2024deepseekmath,jin2025searchr1}.

However, constructing appropriate rewards for search trajectories remains challenging. Directly using the final answer as the reward is easy to verify, but the signal is sparse and ignores correct intermediate queries, evidence selection, and reasoning steps within the trajectory. LLM-as-a-judge provides a scalable way to evaluate intermediate processes that cannot be directly verified~\cite{zheng2023judging}. For example, LeTS combines process-level signals with outcome rewards~\cite{zhang2025lets}, and CW-GRPO further uses an LLM judge to evaluate the reasoning process in each search round~\cite{wang2026enhancing}. Early automatic prompt optimization methods provide a new direction for reducing manual prompt design. APE~\cite{zhou2023ape} treats prompts as candidate instructions and selects the prompt with the highest score according to its performance on the target task. MIPRO~\cite{opsahl2024optimizing} further jointly optimizes instructions and demonstrations in multi-stage language-model programs, while GEPA~\cite{agrawal2025gepa} evolves prompts using feedback from execution trajectories. These methods mainly optimize prompts from the perspective of final task performance and are suitable for improving model outputs. However, for the process reward of a search agent, the goal of the judge prompt is not only to improve the final answer score, but also to stably characterize evaluation dimensions such as query rationality, evidence effectiveness, and reasoning consistency. Based on this observation, judge prompts are generated from the perspective of constraint satisfaction and used for intermediate process scoring in LLM-as-a-judge, providing a complementary perspective to final-score-driven prompt optimization for reinforcement learning of search agents.

To this end, we propose end-to-end automatic prompt engineering (E2E-APE), a constraint-guided method for generating judge prompts in LLM-as-a-judge. Instead of optimizing prompts mainly according to final task performance, E2E-APE starts from the constraints required by intermediate process evaluation and generates judge prompts that focus on query rationality, evidence effectiveness, and reasoning consistency. Furthermore, we introduce EviBack, a hybrid reward framework that combines verifiable final-answer rewards with LLM-judged process rewards. This design enables reinforcement learning to optimize final answers while providing fine-grained feedback on intermediate reasoning in dynamic retrieval.

Our contributions are as follows:

\begin{itemize}
    \item We propose E2E-APE, a constraint-based end-to-end APE method for judge prompt generation. Different from existing prompt optimization methods that mainly rely on final task performance to search prompts, E2E-APE starts from the constraints required by intermediate process evaluation and generates evaluation prompts for LLM-as-a-judge, thereby reducing the cost of manual prompt design and improving the stability of process scoring.

    \item We construct EviBack, a hybrid reward framework for search-agent RAG, which combines verifiable final-answer rewards with Teacher-derived rewards that assess how well Actor trajectories gather supporting evidence and form answers. This design extends reinforcement-learning supervision from final outcomes to the quality of the trajectories that produce them.

    \item Extensive experiments on benchmark datasets show that EviBack improves QA performance over strong baselines, achieving a 16\% gain over baseline.
\end{itemize}

\section{Related Work}

\paragraph{Search Agent.}
Search agents model retrieval in RAG as multi-step interactions, enabling LLMs to generate queries, invoke retrieval tools, and update their reasoning and answers based on the returned evidence~\cite{jin2025searchr1,chen2025research,li2025webthinker}. Recent studies commonly train search agents with reinforcement learning algorithms such as GRPO. However, designing appropriate rewards for trajectories involving multi-round querying, retrieval, and reasoning remains a key challenge. Using only the final answer as the reward is easy to verify, but it provides sparse signals and may overlook correct intermediate steps~\cite{xiong2025raggym}.

To mitigate this issue, LeTS~\cite{zhang2025lets} reflects the intermediate process through indirect signals, while CW-GRPO~\cite{wang2026enhancing} directly evaluates the search process but still relies on manually predefined judge prompts. Unlike these methods, this work studies the automatic generation of judge prompts. Specifically, evaluation prompts are generated based on explicit process constraints, and intermediate process scores are combined with final-answer rewards to construct a hybrid reward for GRPO training.

\paragraph{Automatic Prompt Engineering.}
APE~\cite{zhou2023ape} first casts prompt engineering as an automatic search problem, where large language models propose candidate instructions and select them according to task performance. Subsequent methods, such as MIPRO~\cite{opsahl2024optimizing} and GEPA~\cite{agrawal2025gepa}, further extend this idea to multi-stage language model programs through instruction-demonstration optimization or reflective prompt evolution. These studies mainly emphasize improving validation performance under a given evaluation objective. Our work is complementary: instead of treating prompt design only as score maximization, we study how prompt composition can be constrained to meet multiple requirements for training-time supervision, including answer quality, grounding reliability, stability, and inference cost.

\section{Methodology}

\subsection{Problem Formulation}

For each question $q$ with reference answers $g$, the Actor samples an eight-trajectory group $G=\{\tau_i\}_{i=1}^{8}$. Let $y_i$ denote the answer from trajectory $\tau_i$ and $e_i=\mathrm{EM}(y_i,g)\in\{0,1\}$ its binary outcome reward. EviBack retains the standard Actor rewards when $\max_i e_i=1$ and applies a fallback only when $e_i=0$ for all trajectories.

\subsection{Method Overview}

The method proceeds in four stages: sampling Actor rollout groups; selectively activating the Teacher only for groups in which no trajectory achieves a verifiable exact match; separating gold-blind evidence assessment from gold-aware answer refinement in a two-stage Teacher; and normalizing and downscaling fallback rewards before the GRPO update. Groups containing a verified hit retain their original Actor rewards, and neither a Teacher nor any reference answer is available at inference time. The following sections first define the Teacher and its evidence boundary, then specify the Actor-first reward, and finally describe the automatic construction of the frozen Teacher policy.

\begin{figure*}[t]
  \centering
  \includegraphics[width=0.92\textwidth]{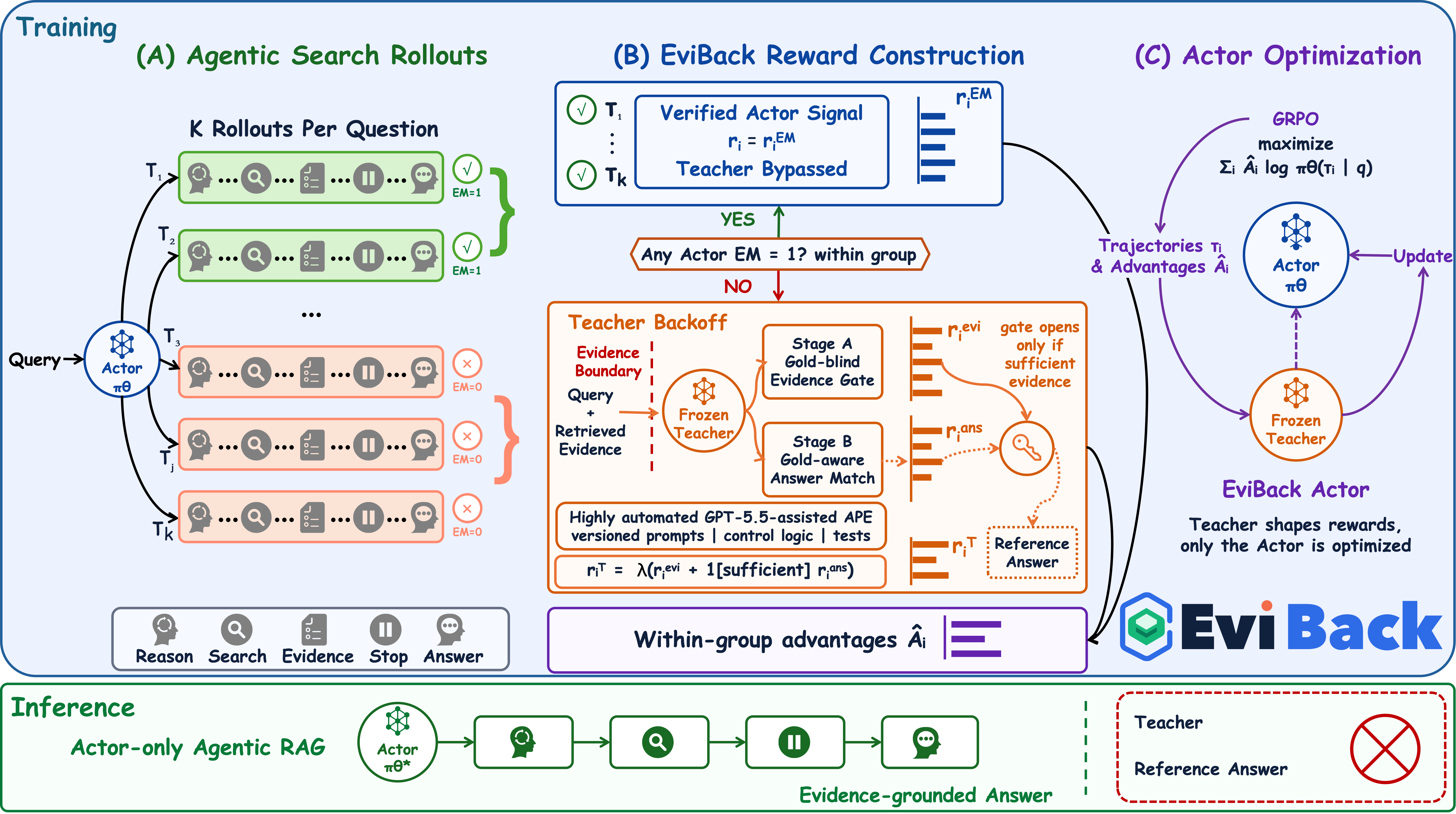}
  \caption{EviBack training and Teacher-free inference. Verified Actor outcomes bypass the Teacher; all-zero groups use evidence-constrained backoff. GPT-5.5 constructs and freezes the Teacher via E2E-APE.}
  \label{fig:framework}
\end{figure*}

\subsection{Reward-Signal Diagnosis}

Across the first eight steps of a Qwen3-1.7B Search-R1 run, 512 eight-rollout groups (4,096 trajectories) yield only 639 exact-match hits (15.6\%): 353 groups are all zero, 21 are all one, and 138 (27.0\%) contain binary contrast. In an all-zero group, the normalized advantage $(e_i-\bar e_G)/(\sigma_G+\epsilon)$ is zero, so partial progress cannot be ranked; mixed groups remain informative relative to verified hits. Group-level backoff is therefore restricted to all-zero groups.

\begin{figure*}[t]
  \centering
  \includegraphics[width=\textwidth]{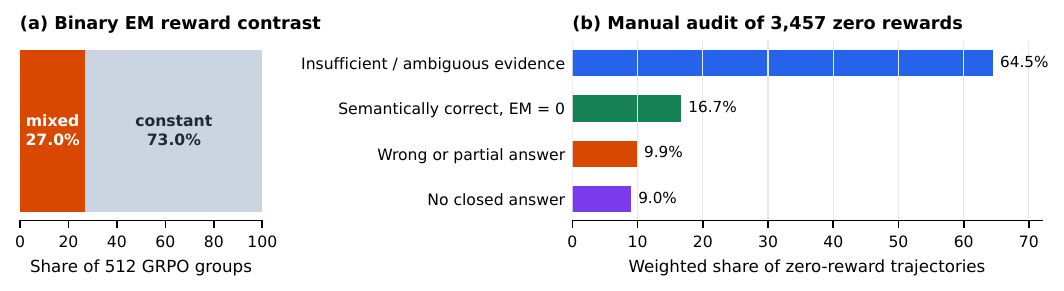}
  \caption{Sparse-reward diagnosis. (a) Binary contrast occurs in 27.0\% of 512 groups. (b) A population-weighted audit partitions zero rewards into evidence failures, answer failures, and exact-match false negatives.}
  \label{fig:sparsity}
\end{figure*}

All 3,457 zero-reward trajectories were stratified by reference-string occurrence and answer closure; 240 were manually labeled using only the question and Actor-visible evidence, with the resulting labels reweighted to the population. Figure~\ref{fig:sparsity} shows that 64.5\% lack sufficient or unambiguous evidence, 16.7\% are semantically correct exact-match false negatives, and 18.9\% have sufficient evidence but a wrong, partial, or unclosed answer. Thus, zero reward conflates failures requiring further search, answer revision, or surface-form correction.

The fallback should therefore be \emph{selective}, \emph{evidence-aware}, and \emph{weak}, preserving mixed-group comparisons and verified Actor outcomes; reference-answer access must not turn incomplete evidence into apparent support.

\subsection{Evidence-Boundary-Preserving Teacher}

The frozen Teacher assesses whether the accumulated Actor-visible evidence $E_i$ is sufficient to answer the original question $q$, not whether the latest search query succeeded. Stage A maps $(q,E_i)$ to a provisional Teacher-reward label in strict XML: a brief rationale, a status $s_i^A\in\{S,I,A\}$, and an evidence-supported answer span when available. It returns supported ($S$) when the evidence supports a complete answer, insufficient ($I$) when any required fact, relation, or reasoning bridge is missing, and ambiguous ($A$) when multiple incompatible complete answers remain possible.

Stage B runs only when $s_i^A\ne I$. Given $(q,E_i,g)$, it calibrates the answer component for reliability and gold alignment. A valid Stage-B output is accepted only if it preserves Stage A's insufficiency boundary; otherwise, the Stage-A output is retained. Thus,
\begin{equation}
  s_i^{\mathrm{final}}=I \quad\Longleftrightarrow\quad s_i^{A}=I.
\end{equation}
A reference-aligned answer form may be selected only when its normalized string occurs in $E_i$. Stage A therefore determines whether a fallback reward is permissible, whereas Stage B improves the quality of the permissible label. Their contributions are asymmetric and complementary rather than duplicate votes.

The override control in Table~\ref{tab:teacher} shows why the gate is necessary. On a 128-case diagnostic set, allowing Stage B to reverse an $I$ decision yielded a Gold-F1 of 0.9222 but reduced insufficient-evidence recall to 0.6993, because an answer string may appear without the relation or reasoning bridge required by the question. Stage B is therefore confined to calibrating the non-$I$ branch.

\subsection{Actor-First Backoff Reward}

For a fallback group, the raw reward for trajectory $i$ is
\begin{equation}
r_i = \beta\,\mathbf{1}[s_i^{\mathrm{final}}\!\ne I]
 + \gamma\,\mathbf{1}[c_i]\,\mathrm{F1}(\hat y_i,g),
\end{equation}
where $\hat y_i$ is the selected Teacher answer, $c_i$ requires a valid closed Actor answer, a valid Teacher output, and a non-$I$ status, and $\beta=\gamma=0.1$. For groups with any successful Actor rollout, $r_i=e_i$ and no Teacher is called.

GRPO standardizes rewards within the group:
\begin{equation}
z_i=\frac{r_i-\mu_G}{\sigma_G+\epsilon},\qquad
A_i=\begin{cases}
z_i,&\max_j e_j=1,\\
\lambda z_i,&\max_j e_j=0,
\end{cases}
\end{equation}
with fallback scale $\lambda=0.1$. Scaling is applied after normalization because scaling every raw fallback reward before standardization would largely cancel. EviBack therefore supplies relative structure to silent groups while keeping their policy-gradient contribution below that of groups with a verifiable hit.

The status term rewards only non-$I$ trajectories, while the answer term ranks them by reference-aligned token F1 only when the Teacher output is valid and the Actor answer is closed. Thus, neither answer scoring nor parser failure can override Stage A's $I$ decision. Ambiguous cases remain on the non-$I$ branch because further search may not resolve an underspecified question; Stage B may canonicalize the answer while preserving the ambiguity in the log.

\subsection{Teacher Construction with E2E-APE}

E2E-APE formulates Teacher construction as constrained policy search rather than manual prompt editing. Before launch, the E2E-APE contract freezes the stratification and $S/I/A$ labeling rules, objectives and metrics, authority constraints, failure behavior, and resource budget. GPT-5.5 first applies this contract to automatically partition raw rollout traces into stratified development and holdout sets and to generate frozen $S/I/A$ labels using only the question and Actor-visible evidence. The manually authored starting point is a single-prompt dual-task Teacher that emits both an evidence-sufficiency judgment and an answer. GPT-5.5 then runs an automatic experiment loop: it generates prompt and workflow candidates, executes cache-free Teacher evaluations, diagnoses metric and trace-level failures in evidence-boundary compliance, stability, and cost, and refines the next hypothesis. The gated separation of evidence assessment from conditional answer refinement is an output of this search. No prompt is edited and no intermediate candidate is selected manually after launch.

\begin{figure}[!t]
  \centering
  \includegraphics[width=\columnwidth]{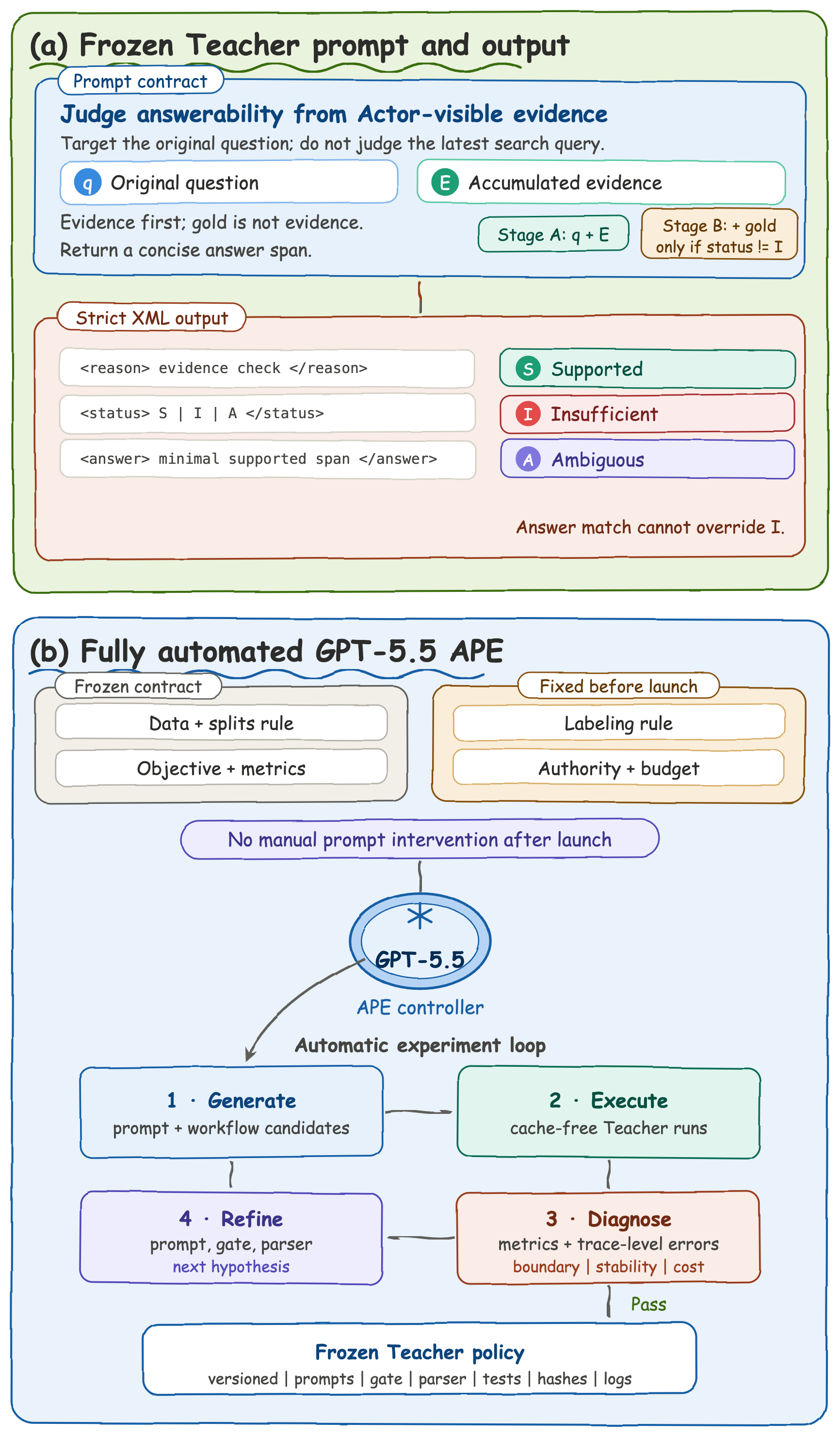}
  \caption{Frozen Teacher interface and E2E-APE construction. (a) Evidence-based status and answer output. (b) GPT-5.5 automatically partitions and labels traces, searches candidates, and freezes the selected Teacher policy.}
  \label{fig:ape}
\end{figure}

Figure~\ref{fig:ape} summarizes both the optimized object and the construction process. Candidates are screened on development data and through repeated cache-free runs; the selected configuration is then frozen and evaluated on held-out data. Selection jointly considers evidence classification, answer coverage, XML validity, repeatability, and inference cost, and admits additional workflow complexity only when it yields stable gains under the predefined budget. The prompts, gate, parser, tests, hashes, and logs are frozen as a versioned Teacher policy. GPT-5.5 serves as the E2E-APE controller, whereas formal RL uses the resulting GLM-4.7-Flash Teacher; neither is present at inference.

\begin{algorithm}[t]
\small
\caption{EviBack training for one question group. Step 2 enforces Actor-first precedence; Step 3c enforces the evidence boundary.}
\label{alg:eviback}
\begin{tabular}{@{}p{0.96\columnwidth}@{}}
\toprule
\textbf{Input:} question $q$, references $g$, Actor $\pi$, group size $N$ \\
1. Sample $\{\tau_i\}_{i=1}^{N}\sim\pi(\cdot\mid q)$ and compute $e_i=\mathrm{EM}(y_i,g)$. \\
2. \textbf{If} $\max_i e_i=1$: set $r_i=e_i$ and skip the Teacher. \\
3. \textbf{Else}, for each $\tau_i$: \\
\quad a. Stage A reads $(q,E_i)$ and returns $(s_i^A,\hat y_i^A)$ without $g$. \\
\quad b. If $s_i^A\ne I$, call Stage B with $(q,E_i,g)$; otherwise stop. \\
\quad c. Select a valid answer, but set $s_i^{\rm final}=I$ iff $s_i^A=I$. \\
\quad d. Compute the bounded status-plus-answer reward $r_i$. \\
4. Normalize $r_i$ within the group; multiply fallback advantages by $\lambda$. \\
5. Apply the GRPO update to $\pi$ and persist rewards, statuses, errors, and call counts. \\
\textbf{Inference:} deploy only $\pi$ with the retriever; omit Teacher and references. \\
\bottomrule
\end{tabular}
\end{algorithm}

\section{Experiments}

The experiments assess answer quality, Teacher-boundary compliance, and search-control costs across Actor scales.

\subsection{Experimental Setup}

\paragraph{Training data and Actors.}
Qwen3 \citep{yang2025qwen3} Actors at 0.6B, 1.7B, and 4B are trained on the same 5,100-example mixture of NQ, HotpotQA, MuSiQue, and 2WikiMultiHopQA, with eight rollouts per prompt. All EviBack settings share the data, seed, reward configuration, and frozen GLM-4.7-Flash Teacher, which is used only during training. Actors follow the Search-R1 interaction protocol.

\paragraph{Retrieval and inference.}
All systems share the Search-R1 wiki-18 corpus and an E5-base-v2 dense retriever \citep{wang2022e5}. A search retrieves 50 candidates and returns the top five passages to the Actor. Actors have at most six assistant turns. Evaluation uses temperature 0, top-$p=1$, full traces, and the same 3,500 examples. The set contains 563 2WikiMultiHopQA \citep{ho2020twowiki}, 125 Bamboogle \citep{press2023bamboogle}, 562 HotpotQA \citep{yang2018hotpotqa}, 562 MuSiQue \citep{trivedi2022musique}, 562 NQ \citep{kwiatkowski2019natural}, 563 PopQA \citep{mallen2023popqa}, and 563 TriviaQA \citep{joshi2017triviaqa} questions.

\paragraph{Baselines.}
The comparison includes the untrained Qwen3 Actor, Search-R1 trained on the same data, and Search-o1 adapted to the shared local retriever; the adapted Search-o1 baseline assesses its control flow rather than its original web-search stack.

\paragraph{Metrics and statistics.}
The evaluation reports legacy EM, token F1, valid-answer rate, mean searches, duplicate-query rate, and max-turn rate. A valid answer is nonempty and properly closed, irrespective of correctness; duplicate-query and max-turn rates record repeated normalized queries and forced termination at six turns. Paired 95\% CIs use 10,000 bootstrap resamples over aligned questions \citep{efron1993bootstrap} and reflect question-level rather than training-seed uncertainty. Dataset-level macros group NQ, PopQA, and TriviaQA as single-hop, and 2Wiki, Bamboogle, HotpotQA, and MuSiQue as multi-hop.

\paragraph{Teacher construction with E2E-APE.}
The frozen GLM-4.7-Flash Teacher is produced by E2E-APE, which automatically partitions and labels traces, evaluates prompt and workflow candidates, and freezes the selected gated policy.

\subsection{Results}

\subsubsection{E2E-APE Teacher Construction and Boundary Audit}

The construction benchmark used by E2E-APE contains 512 trajectories automatically stratified and labeled from the question and Actor-visible evidence, with frozen $S/I/A$ counts of 241/241/30 and a 384/128 development--holdout split. Across 41 recorded experiments, the pipeline generated 12,672 predictions; the selected policy was verified in three cache-free runs and frozen with tests enforcing the non-overridable $I$ gate and conservative failure handling.

\begin{table}[t]
\centering
\small
\setlength{\tabcolsep}{3.8pt}
\begin{tabular}{lrrrr}
\toprule
Teacher policy & I-F1 & Gold-F1 & Objective & Calls \\
\midrule
Gold-blind Stage A & .8924 & .3180 & .6052 & 1.000$\times$ \\
Gold-aware single & .8651 & .6399 & .7525 & 1.000$\times$ \\
\textbf{EviBack gate} & \textbf{.8924} & \textbf{.6825} & \textbf{.7874} & 1.356$\times$ \\
Override control & .8045 & .9222 & .8634 & 1.790$\times$ \\
\bottomrule
\end{tabular}
\caption{Teacher-policy diagnostics. First three rows: three-run means on 221 Teacher-called development cases; override: reused-holdout diagnostic. Objective averages I-F1 and Gold-F1.}
\label{tab:teacher}
\end{table}

Table~\ref{tab:teacher} shows that the EviBack gate preserves Stage A's I-F1 of .8924 while increasing Gold-F1 from .3180 to .6825, exceeding the gold-aware single prompt by .0426. Although the override control scores highest on Gold-F1 and the averaged objective, its insufficient-evidence recall falls to .6993, showing that gold alignment alone cannot guarantee boundary compliance.

During formal 1.7B training, all 40,448 trajectory records preserve the Stage-A boundary. Stage B is called for 37.74\% of Teacher-called trajectories, yielding 1.377 calls per fallback trajectory, close to the 1.356 development estimate and below the two-call budget. Actor EM rises from .192 to .295 as the fallback-group rate falls from .644 to .553; this within-run trend is descriptive.


\subsubsection{Actor Performance and Behavioral Analysis}

\begin{table*}[!t]
\centering
\small
\setlength{\tabcolsep}{5pt}
\begin{tabular}{llrrrrrr}
\toprule
Scale & Method & EM & F1 & Valid ans. & Searches & Duplicate & Max-turn \\
\midrule
0.6B & Base Actor & .0551 & .1337 & .8283 & .7980 & .0003 & .0000 \\
     & Search-o1 & .0694 & .1207 & .8626 & .4580 & .0009 & .0000 \\
     & Search-R1 & .1674 & .2431 & .9966 & .9983 & .0000 & .0000 \\
     & EviBack & \textbf{.1714} & \textbf{.2490} & .9903 & .9986 & .0000 & .0000 \\
\midrule
1.7B & Base Actor & .0749 & .1559 & .6109 & 2.2557 & .3349 & .2046 \\
     & Search-o1 & .1091 & .1762 & .9949 & .6066 & .0037 & .0017 \\
     & Search-R1 & .1800 & .2509 & .7317 & 1.7291 & .1786 & .1549 \\
     & EviBack & \textbf{.2069} & \textbf{.2911} & .8611 & 1.5934 & .1331 & .1071 \\
\midrule
4B   & Base Actor & .2609 & .3720 & .9363 & 1.6837 & .1034 & .0554 \\
     & Search-o1 & .2251 & .3150 & .7911 & 1.2183 & .0289 & .0000 \\
     & Search-R1 & .3023 & .4044 & .8837 & 2.0454 & .1371 & .1160 \\
     & EviBack & \textbf{.3157} & \textbf{.4177} & .8929 & 2.4983 & .2480 & .1063 \\
\bottomrule
\end{tabular}
\caption{Common-protocol results on 3,500 questions. ``Valid ans.'' denotes a nonempty, grammar-closed answer; bold marks the best EM and F1 per scale. Rows are deterministic checkpoint evaluations.}
\label{tab:main}
\end{table*}

\begin{figure*}[!t]
  \centering
  \includegraphics[width=\textwidth]{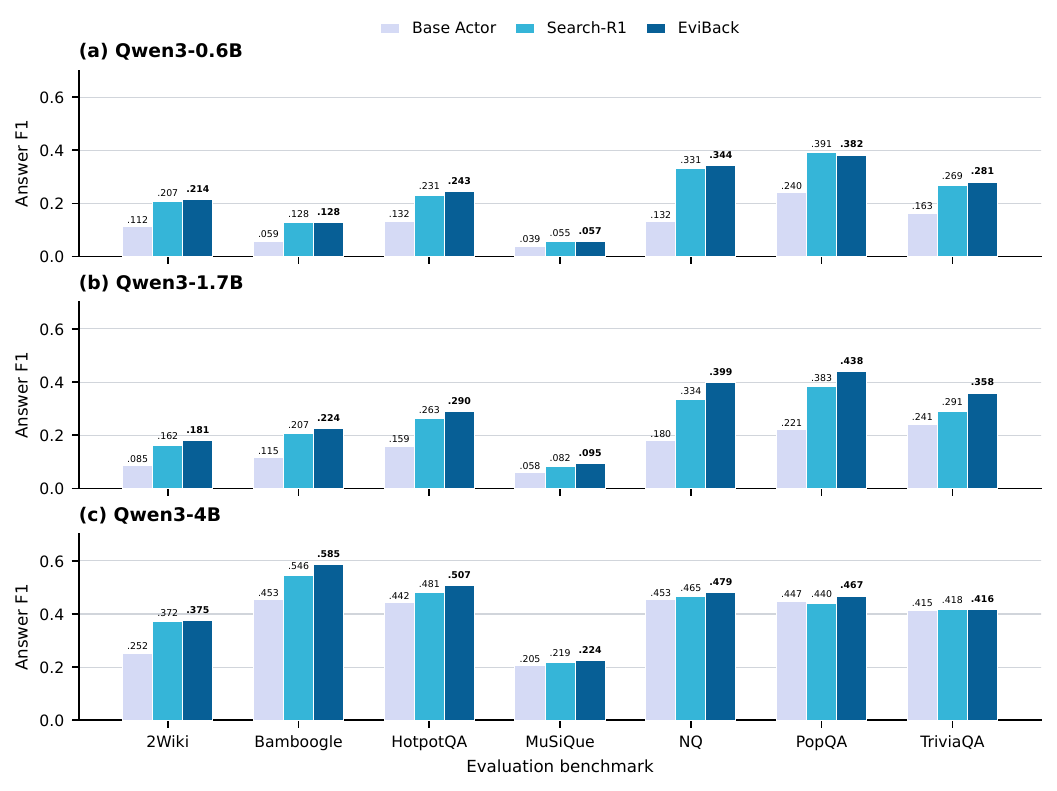}
  \caption{Per-benchmark Answer F1 for the untrained Base Actor, Search-R1, and EviBack across three Qwen3 scales (3,500-question protocol). Bold labels mark EviBack.}
  \label{fig:results}
\end{figure*}

\paragraph{Performance across Actor scales.}
Table~\ref{tab:main} shows that EviBack improves F1 over Search-R1 by $+0.0059$, $+0.0402$, and $+0.0132$ at 0.6B, 1.7B, and 4B, with paired 95\% CIs of $[-0.0015,0.0132]$, $[0.0305,0.0496]$, and $[0.0037,0.0229]$. The gains are clearly positive at 1.7B and 4B, while the 0.6B interval crosses zero. Figure~\ref{fig:results} shows the highest F1 on all seven benchmarks at 1.7B and on six of seven at both 0.6B and 4B; EviBack also exceeds the Base Actor throughout, and improves both single- and multi-hop macro F1 at every scale.

\paragraph{Search and stopping behavior.}
At 1.7B, EviBack raises valid-answer rate from 0.7317 to 0.8611 while reducing mean searches from 1.7291 to 1.5934, duplicate-query rate from 0.1786 to 0.1331, and max-turn termination from 0.1549 to 0.1071. Table~\ref{tab:hop} shows higher single- and multi-hop macro F1 at every scale: the 1.7B gains of 0.0625 and 0.0190 coincide with fewer searches in both categories, whereas the 4B gains of 0.0131 and 0.0185 accompany an increase in mean searches from 2.05 to 2.50 and in duplicate-query rate from 13.7\% to 24.8\%, revealing a scale-dependent control-cost trade-off.

\begin{table}[t]
\centering
\small
\setlength{\tabcolsep}{4pt}
\begin{tabular}{llrrr}
\toprule
Scale & Method & Single F1 & Multi F1 & Search S/M \\
\midrule
0.6B & Search-R1 & .3300 & .1554 & 1.00/1.00 \\
     & EviBack & .3355 & .1606 & 1.00/1.00 \\
1.7B & Search-R1 & .3359 & .1785 & 1.59/1.77 \\
     & EviBack & .3985 & .1975 & 1.49/1.61 \\
4B   & Search-R1 & .4410 & .4045 & 1.69/2.30 \\
     & EviBack & .4541 & .4230 & 2.31/2.58 \\
\bottomrule
\end{tabular}
\caption{Single-/multi-hop macro behavior for trained systems. ``Search S/M'' denotes mean calls in each category.}
\label{tab:hop}
\end{table}

\paragraph{Paired trace analysis.}
Table~\ref{tab:paired} shows that EviBack improves F1 on 354 paired 1.7B questions and lowers it on 156, with 2,990 ties. More importantly, 593 questions recover from \emph{max-turn} or \emph{no-valid-answer} failures to valid answers, versus 140 regressions; their mean F1 changes are $+0.2768$ and $-0.2003$. The gain is therefore concentrated in trajectories that previously failed to emit a valid answer rather than being uniform across already solved questions.

\begin{table}[t]
\centering
\small
\setlength{\tabcolsep}{4pt}
\begin{tabular}{lrr}
\toprule
Paired event & Count & Share \\
\midrule
F1 improves & 354 & 10.1\% \\
F1 declines & 156 & 4.5\% \\
F1 ties & 2,990 & 85.4\% \\
EM improves & 175 & 5.0\% \\
EM declines & 81 & 2.3\% \\
EM ties & 3,244 & 92.7\% \\
Failure $\rightarrow$ valid answer & 593 & 16.9\% \\
Valid answer $\rightarrow$ failure & 140 & 4.0\% \\
\bottomrule
\end{tabular}
\caption{Paired 1.7B events on the same 3,500 questions. Failure combines max-turn and no-valid-answer.}
\label{tab:paired}
\end{table}

\subsection{Ablation Study}

Table~\ref{tab:ablation} evaluates the downstream effects of Teacher construction, fallback strength, and evidence-constrained composition at the 1.7B scale.

\begin{table}[t]
\centering
\scriptsize
\setlength{\tabcolsep}{2.3pt}
\textbf{(a) Answer quality}\\[-2pt]
\begin{tabular}{lcrrr}
\toprule
Variant & $\lambda$ & EM $\uparrow$ & F1 $\uparrow$ & Valid $\uparrow$ \\
\midrule
Base Actor & -- & .0749 & .1559 & .6109 \\
Search-R1 & -- & .1800 & .2509 & .7317 \\
\midrule
Manual Dual-Task$^\dagger$ & -- & .1901 & .2651 & .6414 \\
Single-Stage Teacher & -- & .1994 & .2787 & .8340 \\
EviBack (Two-Stage) & .1 & .2069 & \textbf{.2911} & .8611 \\
EviBack (Two-Stage) & .3 & \textbf{.2080} & \textbf{.2911} & .8203 \\
EviBack (Two-Stage) & .5 & .2049 & .2859 & .8091 \\
EviBack (Two-Stage) & 1.0 & .2043 & .2867 & .7349 \\
\bottomrule
\end{tabular}
\vspace{4pt}

\textbf{(b) Search behavior}\\[-2pt]
\begin{tabular}{lrrr}
\toprule
Variant & Searches $\downarrow$ & Duplicate $\downarrow$ & Max-turn $\downarrow$ \\
\midrule
Base Actor & 2.2557 & .3349 & .2046 \\
Search-R1 & 1.7291 & .1786 & .1549 \\
\midrule
Manual Dual-Task$^\dagger$ & 3.0001 & .5933 & .3279 \\
Single-Stage Teacher & 1.6969 & .1571 & .1369 \\
EviBack (Two-Stage), $\lambda=.1$ & \textbf{1.5934} & \textbf{.1331} & \textbf{.1071} \\
EviBack (Two-Stage), $\lambda=.3$ & 1.9960 & .3186 & .1431 \\
EviBack (Two-Stage), $\lambda=.5$ & 2.4454 & .4049 & .1991 \\
EviBack (Two-Stage), $\lambda=1.0$ & 2.6337 & .5343 & .2320 \\
\bottomrule
\end{tabular}
\caption{Strategy ablations with Qwen3-1.7B on 3,500 questions. The first two rows are reference systems. $^\dagger$Manual single-prompt dual-task control; aggregate metrics only, excluded from paired tests.}
\label{tab:ablation}
\end{table}

\paragraph{Fallback strength.}
Under fixed Hard-Gate v2, values above $\lambda=0.1$ yield no consistent overall gain: $\lambda=0.3$ gives comparable EM and F1, but lowers the valid-answer rate and search efficiency, with similar adverse fluctuations at larger values. Accordingly, $\lambda=0.1$ is retained as the default operating point.

\paragraph{Two-stage evidence constraint.}
Compared with the Single-Stage Teacher, EviBack improves F1 by 0.0124 (paired 95\% CI $[0.0022,0.0223]$) and valid-answer rate by 0.0271 while improving all three search-control metrics, supporting its gated two-stage design.

\paragraph{E2E-APE Teacher construction.}
Relative to the manually authored single-prompt dual-task control (F1=.2651), E2E-APE produces a gated two-stage Teacher through automated decomposition and selection, improving F1 by .0260 and valid-answer rate by .2197 while reducing mean searches by 1.4067 and lowering duplicate-query and forced-termination rates. Since the evidence gate and conditional answer stage are E2E-APE outputs, this comparison measures its downstream contribution to Teacher construction.

\section{Discussion and Limitations}

Teacher LLMs can potentially extract supervision beyond evidence sufficiency and answer alignment, including query quality, evidence coverage, redundancy, and stopping confidence. Experiments with such fine-grained signals did not identify reward constructions that delivered stable gains across settings. EviBack therefore restricts its current reward design to signals with sufficiently consistent empirical utility. Robust reward construction for richer Teacher-derived signals remains an important direction for future work.

Both the training and evaluation sets used in this study are sampled from their respective full datasets. Consequently, the reported results do not directly establish performance on the complete datasets.

\section{Conclusion}

This paper introduced EviBack, a selective Teacher backoff for all-zero rollout groups that separates gold-blind evidence assessment from gold-aware answer alignment, restoring auxiliary supervision without allowing references to override insufficient evidence. The GPT-5.5-assisted E2E-APE pipeline transforms a manual dual-task Teacher into a frozen gated two-stage policy; relative to the manual design, the resulting Teacher improves F1 by .0260 and valid-answer rate by .2197 while reducing mean searches by 1.4067, and remains absent at inference. Across seven QA benchmarks and three Qwen3 scales, EviBack yields higher F1 point estimates than Search-R1 and improves both single- and multi-hop macro F1.

\clearpage
\bibliography{references}

@inproceedings{lewis2020rag,
  title={Retrieval-Augmented Generation for Knowledge-Intensive NLP Tasks},
  author={Lewis, Patrick and Perez, Ethan and Piktus, Aleksandra and Petroni, Fabio and Karpukhin, Vladimir and Goyal, Naman and Kuttler, Heinrich and Lewis, Mike and Yih, Wen-tau and Rocktaschel, Tim and Riedel, Sebastian and Kiela, Douwe},
  booktitle={Advances in Neural Information Processing Systems},
  volume={33},
  pages={9459--9474},
  year={2020}
}

@article{wang2022e5,
  title={Text Embeddings by Weakly-Supervised Contrastive Pre-training},
  author={Wang, Liang and Yang, Nan and Huang, Xiaolong and Jiao, Binxing and Yang, Linjun and Jiang, Daxin and Majumder, Rangan and Wei, Furu},
  journal={arXiv preprint arXiv:2212.03533},
  year={2022}
}

@inproceedings{yao2023react,
  title={{ReAct}: Synergizing Reasoning and Acting in Language Models},
  author={Yao, Shunyu and Zhao, Jeffrey and Yu, Dian and Du, Nan and Shafran, Izhak and Narasimhan, Karthik and Cao, Yuan},
  booktitle={International Conference on Learning Representations},
  year={2023}
}

@inproceedings{asai2024selfrag,
  title={{Self-RAG}: Learning to Retrieve, Generate, and Critique through Self-Reflection},
  author={Asai, Akari and Wu, Zeqiu and Wang, Yizhong and Sil, Avirup and Hajishirzi, Hannaneh},
  booktitle={International Conference on Learning Representations},
  year={2024}
}

@article{jin2025searchr1,
  title={{Search-R1}: Training {LLM}s to Reason and Leverage Search Engines with Reinforcement Learning},
  author={Jin, Bowen and Zeng, Hansi and Yue, Zhenrui and Wang, Dong and Zamani, Hamed and Han, Jiawei},
  journal={arXiv preprint arXiv:2503.09516},
  year={2025}
}

@article{chen2025research,
  title={{ReSearch}: Learning to Reason with Search for {LLM}s via Reinforcement Learning},
  author={Chen, Mingyang and Li, Tianpeng and Sun, Haoze and Zhou, Yijie and Zhu, Chenzheng and Yang, Fan and Zhou, Zenan and Chen, Weipeng and Wang, Haofen and Pan, Jeff Z. and Zhang, Wen and Chen, Huajun},
  journal={arXiv preprint arXiv:2503.19470},
  year={2025}
}

@article{xiong2025raggym,
  title={{RAG-Gym}: Optimizing Reasoning and Search Agents with Process Supervision},
  author={Xiong, Guangzhi and Jin, Qiao and Wang, Xiao and Fang, Yin and Liu, Haolin and Yang, Yifan and Chen, Fangyuan and Song, Zhixing and Wang, Dengyu and Zhang, Minjia and others},
  journal={arXiv preprint arXiv:2502.13957},
  year={2025}
}

@article{shao2024deepseekmath,
  title={{DeepSeekMath}: Pushing the Limits of Mathematical Reasoning in Open Language Models},
  author={Shao, Zhihong and Wang, Peiyi and Zhu, Qihao and Xu, Runxin and Song, Junxiao and Bi, Xiao and Zhang, Haowei and Zhang, Mingchuan and Li, Y. K. and Wu, Y. and Guo, Daya},
  journal={arXiv preprint arXiv:2402.03300},
  year={2024}
}

@article{yang2025qwen3,
  title={{Qwen3} Technical Report},
  author={Yang, An and Li, Anfeng and Yang, Baosong and Zhang, Beichen and Hui, Binyuan and Zheng, Bo and Yu, Bowen and Gao, Chang and Huang, Chengen and Lv, Chenxu and others},
  journal={arXiv preprint arXiv:2505.09388},
  year={2025}
}

@inproceedings{zhou2023ape,
  title={Large Language Models Are Human-Level Prompt Engineers},
  author={Zhou, Yongchao and Muresanu, Andrei Ioan and Han, Ziwen and Paster, Keiran and Pitis, Silviu and Chan, Harris and Ba, Jimmy},
  booktitle={International Conference on Learning Representations},
  year={2023}
}

@article{kwiatkowski2019natural,
  title={Natural Questions: A Benchmark for Question Answering Research},
  author={Kwiatkowski, Tom and Palomaki, Jennimaria and Redfield, Olivia and Collins, Michael and Parikh, Ankur and Alberti, Chris and Epstein, Danielle and Polosukhin, Illia and Devlin, Jacob and Lee, Kenton and others},
  journal={Transactions of the Association for Computational Linguistics},
  volume={7},
  pages={452--466},
  year={2019}
}

@inproceedings{joshi2017triviaqa,
  title={{TriviaQA}: A Large Scale Distantly Supervised Challenge Dataset for Reading Comprehension},
  author={Joshi, Mandar and Choi, Eunsol and Weld, Daniel S. and Zettlemoyer, Luke},
  booktitle={Proceedings of the 55th Annual Meeting of the Association for Computational Linguistics},
  pages={1601--1611},
  year={2017}
}

@inproceedings{mallen2023popqa,
  title={When Not to Trust Language Models: Investigating Effectiveness of Parametric and Non-Parametric Memories},
  author={Mallen, Alex and Asai, Akari and Zhong, Victor and Das, Rajarshi and Khashabi, Daniel and Hajishirzi, Hannaneh},
  booktitle={Proceedings of the 61st Annual Meeting of the Association for Computational Linguistics},
  pages={9802--9822},
  year={2023}
}

@inproceedings{yang2018hotpotqa,
  title={{HotpotQA}: A Dataset for Diverse, Explainable Multi-hop Question Answering},
  author={Yang, Zhilin and Qi, Peng and Zhang, Saizheng and Bengio, Yoshua and Cohen, William W. and Salakhutdinov, Ruslan and Manning, Christopher D.},
  booktitle={Proceedings of the 2018 Conference on Empirical Methods in Natural Language Processing},
  pages={2369--2380},
  year={2018}
}

@inproceedings{ho2020twowiki,
  title={Constructing a Multi-hop {QA} Dataset for Comprehensive Evaluation of Reasoning Steps},
  author={Ho, Xanh and Duong Nguyen, Anh-Khoa and Sugawara, Saku and Aizawa, Akiko},
  booktitle={Proceedings of the 28th International Conference on Computational Linguistics},
  pages={6609--6625},
  year={2020}
}

@article{trivedi2022musique,
  title={{MuSiQue}: Multihop Questions via Single-hop Question Composition},
  author={Trivedi, Harsh and Balasubramanian, Niranjan and Khot, Tushar and Sabharwal, Ashish},
  journal={Transactions of the Association for Computational Linguistics},
  volume={10},
  pages={539--554},
  year={2022}
}

@inproceedings{press2023bamboogle,
  title={Measuring and Narrowing the Compositionality Gap in Language Models},
  author={Press, Ofir and Zhang, Muru and Min, Sewon and Schmidt, Ludwig and Smith, Noah A. and Lewis, Mike},
  booktitle={Findings of the Association for Computational Linguistics: EMNLP 2023},
  pages={5687--5711},
  year={2023}
}

@book{efron1993bootstrap,
  title={An Introduction to the Bootstrap},
  author={Efron, Bradley and Tibshirani, Robert J.},
  publisher={Chapman and Hall/CRC},
  year={1993}
}

@article{openai2023gpt4,
  title={{GPT-4} Technical Report},
  author={{OpenAI}},
  journal={arXiv preprint arXiv:2303.08774},
  year={2023}
}

@article{zhao2023survey,
  title={A Survey of Large Language Models},
  author={Zhao, Wayne Xin and Zhou, Kun and Li, Junyi and Tang, Tianyi and Wang, Xiaolei and Hou, Yupeng and Min, Yingqian and Zhang, Beichen and Zhang, Junjie and Dong, Zican and others},
  journal={arXiv preprint arXiv:2303.18223},
  year={2023}
}

@article{ji2023survey,
  title={Survey of Hallucination in Natural Language Generation},
  author={Ji, Ziwei and Lee, Nayeon and Frieske, Rita and Yu, Tiezheng and Su, Dan and Xu, Yan and Ishii, Etsuko and Bang, Yejin and Chen, Delong and Dai, Wenliang and Chan, Ho Shu and Madotto, Andrea and Fung, Pascale},
  journal={ACM Computing Surveys},
  volume={55},
  number={12},
  pages={1--38},
  year={2023}
}

@article{huang2023survey,
  title={A Survey on Hallucination in Large Language Models: Principles, Taxonomy, Challenges, and Open Questions},
  author={Huang, Lei and Yu, Weijiang and Ma, Weitao and Zhong, Weihong and Feng, Zhangyin and Wang, Haotian and Chen, Qianglong and Peng, Weihua and Feng, Xiaocheng and Qin, Bing and Liu, Ting},
  journal={arXiv preprint arXiv:2311.05232},
  year={2023}
}

@article{singh2026agenticrag,
  title={Agentic Retrieval-Augmented Generation: A Survey on Agentic {RAG}},
  author={Singh, Aditi and Ehtesham, Abul and Kumar, Saket and Khoei, Tala Talaei and Vasilakos, Athanasios V.},
  journal={arXiv preprint arXiv:2501.09136},
  year={2026}
}

@article{gao2023retrieval,
  title={Retrieval-Augmented Generation for Large Language Models: A Survey},
  author={Gao, Yunfan and Xiong, Yun and Gao, Xinyu and Jia, Kangxiang and Pan, Jinliu and Bi, Yuxi and Dai, Yi and Sun, Jiawei and Wang, Meng and Wang, Haofen},
  journal={arXiv preprint arXiv:2312.10997},
  year={2023}
}

@article{li2025webthinker,
  title={{WebThinker}: Empowering Large Reasoning Models with Deep Research Capability},
  author={Li, Xiaoxi and Jin, Jiajie and Dong, Guanting and Qian, Hongjin and Wu, Yongkang and Wen, Ji-Rong and Zhu, Yutao and Dou, Zhicheng},
  journal={arXiv preprint arXiv:2504.21776},
  year={2025}
}

@article{zheng2023judging,
  title={Judging {LLM}-as-a-Judge with {MT-Bench} and Chatbot Arena},
  author={Zheng, Lianmin and Chiang, Wei-Lin and Sheng, Ying and Zhuang, Siyuan and Wu, Zhanghao and Zhuang, Yonghao and Lin, Zi and Li, Zhuohan and Li, Dacheng and Xing, Eric P. and Zhang, Hao and Gonzalez, Joseph E. and Stoica, Ion},
  journal={arXiv preprint arXiv:2306.05685},
  year={2023}
}

@article{zhang2025lets,
  title={{LeTS}: Learning to Think-and-Search via Process-and-Outcome Reward Hybridization},
  author={Zhang, Qi and Yang, Shouqing and Gao, Lirong and Chen, Hao and Hu, Xiaomeng and Chen, Jinglei and Wang, Jiexiang and Guo, Sheng and Zheng, Bo and Wang, Haobo and Zhao, Junbo},
  journal={arXiv preprint arXiv:2505.17447},
  year={2025}
}

@article{wang2026enhancing,
  title={Enhancing {LLM}-based Search Agents via Contribution Weighted Group Relative Policy Optimization},
  author={Wang, Junzhe and Xi, Zhiheng and Yang, Yajie and Luo, Hao and Dou, Shihan and Gui, Tao and Zhang, Qi},
  journal={arXiv preprint arXiv:2604.14267},
  year={2026}
}

@inproceedings{opsahl2024optimizing,
  title={Optimizing Instructions and Demonstrations for Multi-Stage Language Model Programs},
  author={Opsahl-Ong, Krista and Ryan, Michael J. and Purtell, Josh and Broman, David and Potts, Christopher and Zaharia, Matei and Khattab, Omar},
  booktitle={Proceedings of the 2024 Conference on Empirical Methods in Natural Language Processing},
  pages={9340--9366},
  year={2024}
}

@article{agrawal2025gepa,
  title={{GEPA}: Reflective Prompt Evolution Can Outperform Reinforcement Learning},
  author={Agrawal, Lakshya A. and Tan, Shangyin and Soylu, Dilara and Ziems, Noah and Khare, Rishi and Opsahl-Ong, Krista and Singhvi, Arnav and Shandilya, Herumb and Ryan, Michael J. and Jiang, Meng and Potts, Christopher and Sen, Koushik and Dimakis, Alexandros G. and Stoica, Ion and Klein, Dan and Zaharia, Matei and Khattab, Omar},
  journal={arXiv preprint arXiv:2507.19457},
  year={2025}
}

\end{document}